\documentclass{article}

\usepackage{times}
\usepackage{graphicx} 

\usepackage{natbib}

\usepackage{algorithm}
\usepackage{algorithmic}

\usepackage{amsmath,amsthm,amsfonts,amssymb}
\usepackage{mathtools} 
\usepackage{bm}

\usepackage{hyperref}
\usepackage{threeparttable}

\newcommand{\eat}[1]{}

\newcommand{\norm}[1]{\left\lVert#1\right\rVert}

\DeclarePairedDelimiter{\<}{<}{>}

\DeclarePairedDelimiter{\abs}{|}{|}

\DeclareMathOperator*{\argmax}{arg\,max}
\numberwithin{equation}{section}

\usepackage[accepted]{icml2014}

\icmltitlerunning{Robotic Search \& Rescue via Online Multi-task Reinforcement Learning}

\begin{document} 

\twocolumn[
\icmltitle{Robotic Search \& Rescue via Online Multi-task Reinforcement Learning}

\icmlauthor{Lisa Lee}{lilee@princeton.edu}
\icmladdress{Department of Mathematics, Princeton University, Princeton, NJ 08544, USA}
\icmlauthor{Advisor: Dr. Eric Eaton}{eeaton@seas.upenn.edu}
\icmlauthor{Mentors: Dr. Haitham Bou Ammar, Christopher Clingerman}{}
\icmladdress{GRASP Laboratory, University of Pennsylvania, Philadelphia, PA 19104, USA}

\icmlkeywords{machine learning, multi-task reinforcement learning, lifelong learning, ELLA, GRASP}

\vskip 0.3in
]

\begin{abstract}
Reinforcement learning (RL) is a general and well-known method that a robot can use to learn an optimal control policy to solve a particular task. We would like to build a versatile robot that can learn multiple tasks, but using RL for each of them would be prohibitively expensive in terms of both time and wear-and-tear on the robot. To remedy this problem, we use the Policy Gradient Efficient Lifelong Learning Algorithm (PG-ELLA), an online multi-task RL algorithm that enables the robot to efficiently learn multiple consecutive tasks by sharing knowledge between these tasks to accelerate learning and improve performance. We implemented and evaluated three RL methods---Q-learning, policy gradient RL, and PG-ELLA---on a ground robot whose task is to find a target object in an environment under different surface conditions. In this paper, we discuss our implementations as well as present an empirical analysis of their learning performance.
\end{abstract}

\section{Introduction}
In the robotic search \& rescue problem (S\&R), we have a robot whose task is to search and rescue hidden target objects in the environment. Reinforcement learning (RL) is one method the robot can utilize to learn an optimal action selection rule, referred to as an optimal policy. Unfortunately, due to large state spaces that are common in robotics, the learning time and memory requirements can quickly become prohibitively expensive as the robot's task becomes more complex. The robot may run out of time or experience mechanical failure before learning is complete, thereby making RL often impractical. Moreover, when the agent is given a new task in a different environment, it must re-learn a new policy from scratch as it loses the knowledge it gained from previous experiences.

Multi-task learning (MTL) helps to ameliorate this problem by enabling the agent to use its experience from previously learned tasks to improve learning performance in a related but different task \cite{caruana97}. In batch MTL, multiple tasks are learned simultaneously, while in online MTL, the tasks are learned consecutively. The latter method is of more interest for S\&R applications, as it allows the agent to continually build on its knowledge over a sequence of tasks, thereby lessening the learning cost for every new task. Using MTL, we can train the robot in simulation ahead of time over a series of tasks, then transfer the learned knowledge to the real robot. The end result is an intelligent, flexible machine which can quickly adapt to new scenarios using learned knowledge, and which can learn new tasks more efficiently \cite{Thrun96b}. 

A robot capable of learning new tasks quickly by building upon previously learned knowledge would be vastly useful in applications such as S\&R, where time constraints are an issue and where the robot is often presented with similar tasks in various scenarios with different environment dynamics.

In this project, we framed the problem as a reinforcement learning (RL) problem, where the agent's task is to find the target object in an environment. To allow variations in the tasks, we changed the coefficient $\bm{\mu}_t$ of ground friction for each task $t$. This is a simplified version of the S\&R problem, focusing on having the robot learn across different scenarios (in this case, ground friction). This setup is applicable in situations where the robot has to drive over various types of grounds, such as rough carpet or sand.

Using ROS and MATLAB, we began by implementing two single-task RL methods on a Turtlebot 2: Q-learning and policy gradient RL (PG). Then we implemented the Policy Gradient Efficient Lifelong Learning Algorithm (PG-ELLA), an online MTL algorithm using PG that allows efficient sharing of knowledge between consecutive tasks to accelerate learning \cite{pgella14}.

Results and analysis of these implementations can be found in Section \ref{testing_and_analysis}. In particular, we discuss experimental results of using a user policy to improve the learning performance of Q-learning, and demonstrate how increasing the size of the state space can exacerbate the learning cost for Q-learning. We also compare the learning speed of PG and Q-learning, and show that the robot learns noticeably faster using PG.

This project marks the first time that the Efficient Lifelong Learning Algorithm (ELLA) \cite{ella13} was implemented on a real robot. PG-ELLA has been proven to be effective for simple dynamical systems \cite{pgella14}, and we created several demonstration videos showing its effectiveness. However due to time constraints, we have not done complete empirical analysis of our PG-ELLA implementation, instead leaving this to future work.

\section{Single-Task Reinforcement Learning}

The reinforcement learning problem can be formulated as a Markov decision process (MDP), where the problem model  $\<*{S, A, P, R,, \gamma}$ consists of a state space $S$, a set of actions $A$, a state transition probability function $P: S \times A \times S \mapsto [0,1]$ describing the environment's dynamics, a reward function $r: S \times A \times S \mapsto \mathbb{R}$, and a discount factor $\gamma \in [0, 1]$ which trades off the importance of sooner versus later rewards. We have an agent and an environment that interact continually, the agent selecting actions and the environment responding to those actions and presenting new situations to the agent. At each time step $t \in \mathbb{Z}^+$, the agent receives the environment's current state $\bm{s}_t \in S$, and on that basis selects an action $\bm{a}_t \in A(\bm{s}_t)$, where $A(\bm{s}_t)$ is the set of legal actions in state $\bm{s}_t$. One time step later, as a consequence of the agent's action and the current state, the environment sends a numerical reward $r(\bm{s}_t, \bm{a}_t, \bm{s}_{t+1}) \in \mathbb{R}$ and a new state $\bm{s}_{t+1}$ to the agent. The agent's goal is to maximize the total expected discounted return, defined as
    \[ R_t = \sum_{k=0}^\infty \gamma^k r_{t+k+1}\enspace , \]
where $r_t := r(\bm{s}_t, \bm{a}_t, \bm{s}_{t+1})$.

\begin{figure}[t]
\vskip 0.2in
\begin{center}
\centerline{\includegraphics[width=0.7\columnwidth]{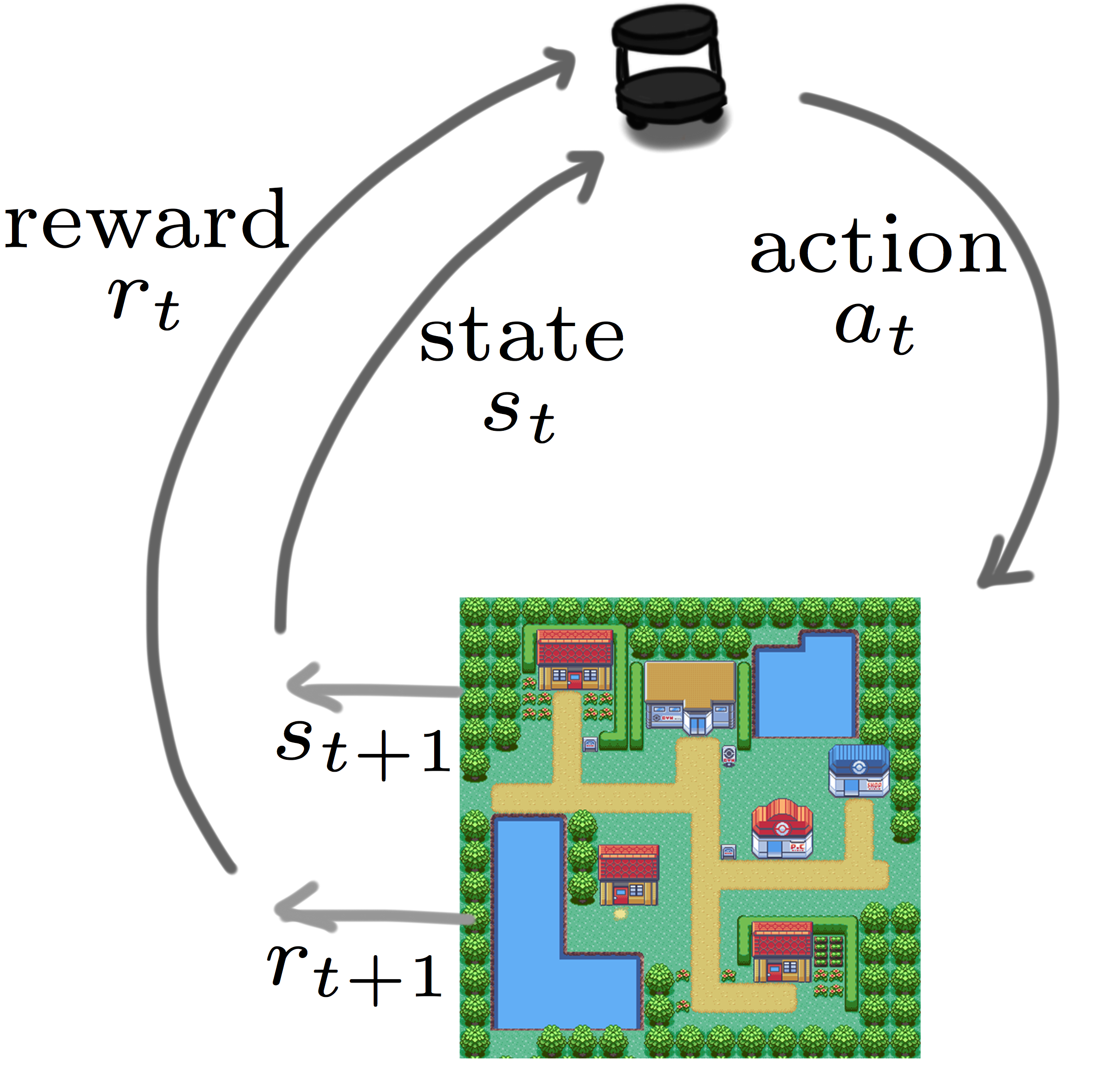}}
\caption{The agent and the environment interacting continually in the reinforcement learning setting.}
\end{center}
\vskip -0.2in
\end{figure}

At each time step, the agent updates its policy $\pi_t: S \times A \mapsto [0, 1]$, where for each $(s, a) \in S \times A$, $\pi_t(s, a)$ is the probability that the agent selects $a_t=a$ if $s_t = s$. Reinforcement learning techniques specify how the agent implements its policy as a result of its experience.  Two RL techniques that we implemented are Q-learning and policy gradient RL (PG), which we briefly desribe below.

    \subsection{Q-learning}
    
    In Q-learning, we have a finite discrete state space $S$ and a finite set of discrete actions $A$. The algorithm stores a function $Q:S\times A \mapsto \mathbb{R}$ which estimates the quality of a state-action pair. Before learning starts, $Q$ is initialized to return an arbitrary fixed value. Then in every time step $t \in \mathbb{Z}^+$, the algorithm chooses an action $a_t = \argmax_{a} Q_t(s_t, a)$ with the highest $Q$-value, then updates its estimate of $Q$ via a simple value iteration update,
 	\begin{align*}
    Q_{t+1}(s_t, a_t)
    &= (1-\alpha)\underbrace{Q_t(s_t, a_t)}_\text{old value}\\
    &+ \alpha \left[\underbrace{r_{t+1}}_\text{reward}
    + \underbrace{\gamma}_{\substack{\text{discount}\\\text{factor}}} \underbrace{\max_a Q(s_{t+1}, a)}_{\substack{\text{estimate of optimal}\\\text{future value}}} \right] \enspace ,
    \end{align*}
where $\alpha \in (0, 1]$ is the learning rate and $\gamma \in [0, 1]$ is the discount factor.

While Q-learning works with discrete states and actions, in robotics it is preferable to work with continuous states and actions, as a robot's action parameters and sensor readings often take on continuous values. Therefore, a more useful RL technique for our purposes is the policy gradient RL (PG), which allows learning with continuous states and actions.

    \subsection{Policy Gradient Reinforcement Learning}\label{subsection:pg}
    
    In policy gradient RL (PG), we have a (possibly infinite) set of states $S \subseteq \mathbb{R}^d$ and a set of actions $A \subseteq \mathbb{R}^m$. The general goal is to optimize the \emph{expected return} of the policy $\pi$ with parameters $\bm{\theta} \in \mathbb{R}^d$, defined by
	    \begin{align*}
	    \mathcal{J}(\bm{\theta}) &= \int_{\mathbb{T}} p_{\bm{\theta}}(\bm{\tau}) R(\bm{\tau}) d\bm{\tau}
	    \end{align*}
    where $\mathbb{T}$ is the set of all possible \emph{trajectories}. A trajectory $\bm{\tau} = [\bm{s}_{1:T+1}, \bm{a}_{1:T}]$ denotes a series of states $\bm{s}_{1:H+1} = [\bm{s}_1, \bm{s}_2, \ldots, \bm{s}_{H+1}]$ and actions $\bm{a}_{1:H} = [\bm{a}_1, \bm{a}_2, \ldots, \bm{a}_H]$. The probability of a trajectory $\bm{\tau}$ given  policy $\bm{\theta}$ is denoted by $p_\theta(\bm{\tau})$, and its average per time step return $R(\tau)$ is
	    \[ R(\bm{\tau}) = {1 \over H} \sum_{t=1}^H r(\bm{s}_t, \bm{a}_t, \bm{s}_{t+1}). \]

    Using the Markov assumption, we can compute the probability of a trajectory as a product of probabilities of states and actions:
	    \begin{align*}
	    p_{\bm{\theta}}(\bm{\tau})
	    = p(\bm{s}_1) \prod_{t=1}^H p(\bm{s}_{t+1} \mid \bm{s}_t, \bm{a}_t) \pi(\bm{a}_t \mid \bm{s}_t, t) \enspace ,
	    \end{align*}
    where $p(\bm{s}_1)$ denotes the initial state distribution, and $p(\bm{s}_{t+1} \mid \bm{s}_t, \bm{a}_t)$ denotes the next state distribution conditioned on the last state and action.

    The next action is sampled from a normal distribution
    \begin{align*}
    \bm{a}_{t+1} \sim \mathcal{N}(\bm{\theta}_t^T \phi(\bm{s}_t), \sigma_t) \enspace ,
    \end{align*}
    where $\phi: S \mapsto \mathbb{R}^d$ is a feature vector. The value iteration update is given by
    \begin{align*}
    \bm{\theta}_{t+1} &= \bm{\theta}_t + \alpha \nabla \\
    \sigma_{t+1} &= \sigma_{t} + \alpha \sigma_t \nabla \enspace ,
    \end{align*}
    where $\alpha \in (0, 1]$ is the learning rate and $\nabla := \nabla_{\bm{\theta}} \mathcal{J}(\bm{\theta})$. For a derivation of the value iteration update, we refer the reader to Kober \& Peters \yrcite{Kober08}.


\section{Multi-Task Reinforcement Learning}

Often in robotics, single-task RL methods suffer from slow convergence speed due to the huge size of the state space. Online MTL provides a solution to this problem by allowing knowledge to be shared between multiple tasks. In the following sections we present the Policy Gradient Efficient Lifelong Learning Algorithm (PG-ELLA), an online MTL algorithm that extends PG to enable a computationally efficient sharing of knowledge between consecutive tasks \cite{pgella14}. PG-ELLA is based on the Efficient Lifelong Learning Algorithm (ELLA), an online MTL algorithm that has proven to be effective in the supervised MTL scenario \cite{ella13}.

\subsection{The Online MTL Problem}

In the online MTL setting, the agent is presented with a series of tasks $\mathcal{Z}^{(1)}, \ldots, \mathcal{Z}^{(T_{\textrm{max}})}$, where each task $t$ is an MDP $\<*{S^{(t)}, A^{(t)}, P^{(t)}, R^{(t)}, \gamma^{(t)}}$ with initial state distribution $P_0^{(t)}$. The agent learns the tasks consecutively, training on multiple trajectories for each task before moving on to the next. We assume that the agent does not know the task order, their distribution, or the total number of tasks.

The agent's goal is to find a set of optimal policies $\bm{\Pi}^* = \{ \pi_{\bm{\theta}^{(1)}}^*, \ldots, \pi_{\bm{\theta}^{(T_{\textrm{max}})}}^*\}$ with corresponding parameters $\bm{\Theta}^* = \{\bm{\theta}^{(1)\ast}, \ldots, \bm{\theta}^{(T_{\textrm{max}})\ast} \}$. The agent aims to optimize its learned policies for all tasks $\mathcal{Z}^{(1)}, \ldots, \mathcal{Z}^{(T)}$, where $T$ is the number of tasks seen so far.

\subsection{PG-ELLA}

As in Section \ref{subsection:pg}, we let $d$ be the dimension of the state space, i.e., $S^{(t)} \subseteq \mathbb{R}^d$ for each task $t$. PG-ELLA extends the PG algorithm to use a shared latent basis $\bm{L} \in \mathbb{R}^{d \times k}$ (where $k$ is the number of latent components) that enables transfer of knowledge between multiple tasks. We assume that the control parameters $\bm{\theta}^{(t)}$ ($t \in \{1, \ldots, T_{\textrm{max}}\}$) can be modeled as a linear combination of the columns of $\bm{L}$, thereby allowing us to write $\bm{\theta}^{(t)} = \bm{L}\bm{s}^{(t)}$, where $\bm{s}^{(t)} \in \mathbb{R}^k$ is a vector of coefficients. 

The aim of the algorithm can be represented by the following objective function,
    \begin{equation}\label{eq:pg_ella}
    \!\!\min_{\bm{L}} \frac{1}{T} \!\sum_{t=1}^T \min_{\bm{s}^{(t)}}
    \!\!\left[ \!
    \underbrace{
        -\mathcal{J}\left(\bm{\theta}^{(t)}\right)
    }_{\substack{\text{model fit to}\\\text{individual task}}}
    \!+\! \underbrace{\mu \norm{\bm{s}^{(t)}}_1}_\text{sparsity}
    \!\right] \!
    +\!\underbrace{
        \lambda \norm{\bm{L}}^2_\text{F}
    }_{\substack{\text{maximize}\\\text{shared}\\\text{knowledge}}} \enspace ,
    \end{equation}
where $\bm{\theta}^{(t)} = \bm{L}\bm{s}^{(t)}$, $\norm{\cdot}_1$ is the $L_1$ norm, and $\norm{\cdot}_\text{F}$ is the Frobenius norm. Just as in PG, we maximize the expected return $\mathcal{J}\left(\bm{\theta}^{(t)}\right)$ (or equivalently, minimize $-\mathcal{J}\left(\bm{\theta}^{(t)}\right)$) of the policy $\pi_{\bm{\theta}^{(t)}}$, for each task $t$. We also minimize the sparsity $\norm{\bm{s}^{(t)}}_1$ of the coefficient vectors $\bm{s}^{(t)}$ for each task $t$, to ensure that each latent basis component encodes a maximal reusable piece of knowledge. Lastly, we minimize the complexity $\norm{\bm{L}}^2_\text{F}$ of the shared latent basis $\bm{L}$, in order to maximize the shared knowledge among tasks.

There are essentially two main steps to the algorithm, for every task $t$: first, we fit the model to the individual task by minizing the term $-\mathcal{J}\left(\bm{\theta}^{(t)}\right) + \mu \norm{\bm{s}^{(t)}}_1$ with respect to $\bm{s}^{(t)}$; then we maximize the shared knowledge across all tasks by minimizing $\lambda \norm{\bm{L}}^2_\text{F}$. For the exact PG-ELLA algorithm and other details, please see Bou Ammar et al. \yrcite{pgella14}.


\section{Methods}

We implemented Q-Learning, PG, and PG-ELLA each as a separate ROS package, and trained the robot in simulation using Gazebo.\footnote{\texttt{gazebosim.org}}. For PG and PG-ELLA, we also used the Policy Gradient Toolbox,\footnote{\texttt{ias.tu-darmstadt.de/Research\\$\phantom{xxx}$/PolicyGradientToolbox}} a MATLAB toolbox for policy gradient methods; and the MATLAB ROS I/O Package,\footnote{\texttt{mathworks.com/ros}} which allows MATLAB to exchange messages with other nodes on the ROS network.

\begin{figure}[t]
\vskip 0.2in
\begin{center}
\centerline{\includegraphics[width=0.7\columnwidth]{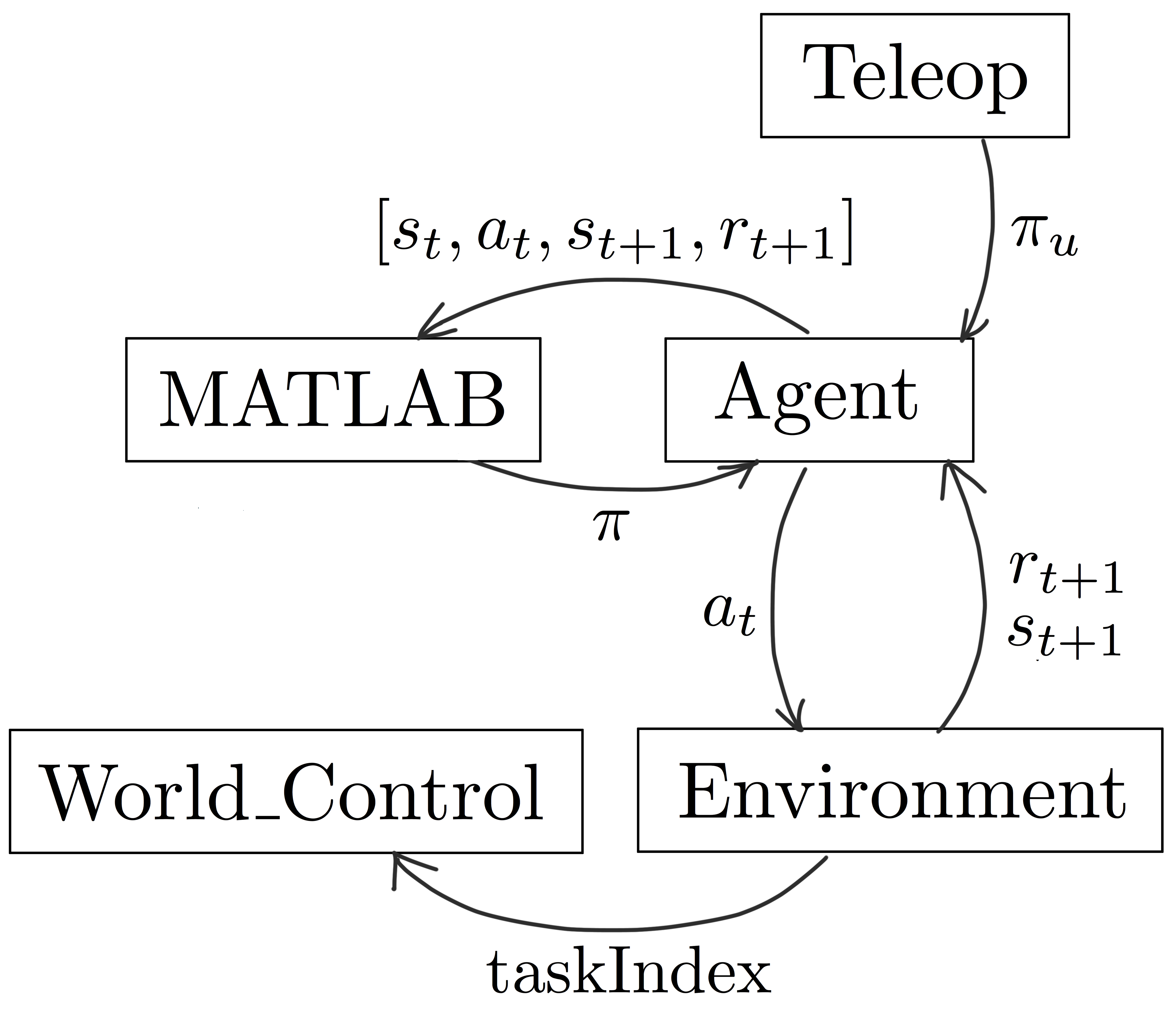}}
\caption{How the ROS nodes in our RL implementations communicate with each other. \texttt{Agent} computes the robot's policy, and sends its action $a_t$ to \texttt{Environment}. Then \texttt{Environment} receives and executes $a_t$, and sends the next state $s_{t+1}$ and reward $r_{t+1}$ to \texttt{Agent}. Every time there is a new task, \texttt{Environment} sends the taskIndex to \texttt{World\_Control}, which then changes the environment dynamics accordingly. For PG and PG-ELLA, \texttt{Agent} sends trajectories $\{ [s_t, a_t, s_{t+1}, r_{t+1}] \}$ to \texttt{MATLAB}, which uses it to update the robot's policy $\pi$. Instead of using $\pi$ to select the next action, \texttt{Agent} can also use the user policy $\pi_u$ (which is given by the user via keyboard \texttt{Teleop}). \label{fig_node}}
\end{center}
\vskip -0.2in
\end{figure}

For each of the three RL packages, we created four nodes---\texttt{Agent}, \texttt{Environment}, \texttt{World\_Control}, and \texttt{Teleop}---that communicate with each other using ROS messages. For PG and PG-ELLA, we also have a fifth note---\texttt{MATLAB}---that receives trajectories from \texttt{Agent} and compute the robot's policy using the episodic REINFORCE policy gradient algorithm \cite{Williams92}. A diagram is shown in Fig \ref{fig_node}. 

    \subsection{Representation}
        \subsubsection{State space}\label{subsubsection:state_space}

\begin{figure}[t]
\vskip 0.2in
\begin{center}
\centerline{\includegraphics[width=0.45\columnwidth]{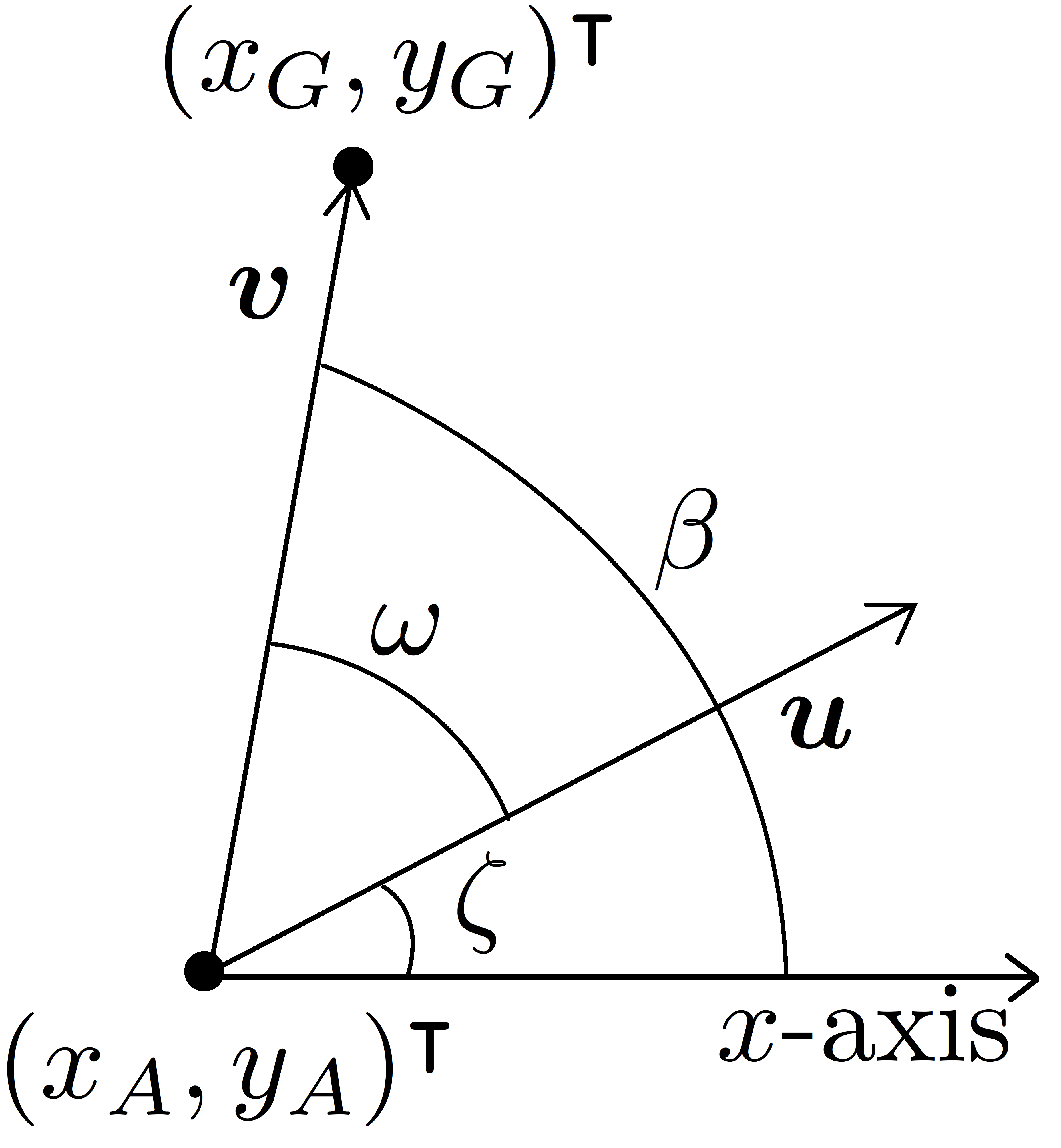}}
\caption{For PG and PG-ELLA, our state representation is $\{ (d, \omega)^\intercal \}$ where $d = \norm{\bm{v}}$. For Q-learning, we use $\{ ([d], [w]/12)^\intercal \}$ or $\{ ([d], [\beta]/12, [\zeta]/12)^\intercal \}$ (see Section \ref{subsubsection:state_space_size}). \label{fig_staterep}}
\end{center}
\vskip -0.2in
\end{figure}

        Let $(x_A, y_A)^\intercal$ and $(x_G, y_G)^\intercal$ be the locations of the agent and of the target object, respectively, on a 2D map of the environment. Let $\bm{v} \in \mathbb{R}^2$ be the vector whose initial and terminal points are $(x_A, y_A)^\intercal$ and $(x_G, y_G)^\intercal$, respectively. Let $\bm{u} \in \mathbb{R}^2$ be a vector whose initial point is $(x_A, y_A)^\intercal$ and is parallel to the agent's orientation. Then for PG and PG-ELLA, we used a state space given by $\{(d, \omega)^\intercal\}$, where $d = \norm{\bm{v}}$ is the distance (in meters) from the agent to the target object, and $\omega$ is the counterclockwise angle (in degrees) from $\bm{u}$ to $\bm{v}$. (See Fig. \ref{fig_staterep}.)
        
        For Q-learning, we discretized the state space to be $\{([d], [\omega]/12)^\intercal\}$, where $[x]$ rounds $x$ to the nearest integer, for any $x \in \mathbb{R}$; and $a/b \in \mathbb{Z}^+$ is the quotient of $a$ divided by $b$, for any $a, b \in \mathbb{Z}$. We took the quotient $[\omega]/12$ (which takes on values in $\{1, \ldots, 30\}$) rather than $[\omega]$, in order to reduce the size of the state space.
        
        It is reasonable to define our state space in this way, as there are S\&R scenarios where the robot has already located the target object, and knows the GPS coordinates of itself and of the target object.

        \subsubsection{Action space} 

For PG and PG-ELLA, the action space is $\{(v_\ell, v_\omega)^\intercal: v_\ell, v_\omega \in \mathbb{R} \}$, where $v_\ell$ is the robot's linear velocity (in meters/second) and $v_\omega$ is the robot's angular velocity (in radians/second). In order to prevent abnormal motor behavior, in cases where the robot learns a bizarre policy, we restrict the range of $v_\ell$ and $v_\omega$ to $[-1.5, 1.5]$. In other words, if the robot's policy returns a value for $v_\ell$ (or $v_\omega$) that exceeds or is below the fixed range, then the robot simply uses the maximum (1.5) or minimum (-1.5) value, respectively.

For Q-learning, which requires a discrete action space, we chose the actions to be simply \{Forward, Left, Right\}.
	
	\subsection{Reward function}
    
    We defined the reward function $r: S \times A \times S \mapsto \mathbb{R}$ in a way such that the agent is penalized for taking too many steps to reach the goal, and is rewarded for finding the target object. More specifically, the reward function we chose for Q-learning is
    \[
    r(s_t, a_t, s_{t+1}) =
    \begin{cases}
    100 & \textrm{if $[d]< 1$ and $[\omega]/12 < 2$}\\
    -1 & \textrm{otherwise} \enspace ,
    \end{cases}
    \]
    where $([d], [\omega]/12)^\intercal= s_{t+1}$.
    Similarly, the reward function we chose for PG and PG-ELLA is
    \[
    r(\bm{s}_t, \bm{a}_t, \bm{s}_{t+1}) =
    \begin{cases}
    100 & \textrm{if $d < 0.55$ and $\omega < 0.2$} \\
    -0.5 \abs{v_\ell} & \textrm{otherwise}  \enspace ,
    \end{cases}
    \]
    where $(d, \omega)^\intercal = \bm{s}_{t+1}$ and $v_\ell$ is the linear velocity of $\bm{a}_t$.

    \subsection{Tasks}
    
    To allow variations in the tasks, we changed the ground friction coefficient $\bm{\mu}_t$ for each task $t$. Ideally we would like to draw the friction coefficients from a normal distribution for each task, but due to time constraints, we instead used fixed friction coefficients for five different tasks (see Table \ref{table:pgtable}).

\noindent\resizebox{\columnwidth}{!}{%
\begin{threeparttable}[t]
\caption{The learning rates $(\alpha_\ell^{(t)}, \alpha_\omega^{(t)})$ chosen for PG using the method described in Section \ref{subsubsection:learning_rates}. $\bm{\mu}_t$ is the ground friction coefficient\textsuperscript{\textdagger} for task $t$. \label{table:pgtable}}
\vskip 0.15in
\begin{center}
\begin{small}
\begin{sc}
\begin{tabular}{lcccr}
\hline
\abovespace\belowspace
Task $t$ & $\bm{\mu}_t$& $(\alpha_\ell^{(t)}, \alpha_\omega^{(t)})$ & Avg. cumul. reward\textsuperscript{\textdagger\textdagger}\\
\hline
\abovespace
1
&(100, 50)& $(10^{-6}, 10^{-7})$ & 77.84 \tabularnewline \\
2
&(5, 5)& $(10^{-6}, 10^{-6})$ & 87.60 \tabularnewline \\
3
&(10, 0.1)& $(10^{-7}, 10^{-5})$ & 51.12 \tabularnewline \\
4
&(0.1, 50)& $(10^{-7}, 10^{-6})$ & 77.32 \tabularnewline \\
5
&(0.2, 0.2) & $(10^{-7}, 10^{-5})$ & 78.32 \tabularnewline \\
\hline
\end{tabular}
\begin{tablenotes}
    \item[\textdagger] \scriptsize{In Gazebo, friction is defined as a 2-dimensional vector $(\mu_1, \mu_2)$ where $\mu_1, \mu_2 \in \mathbb{R}^+$. (See {\tt gazebosim.org/wiki/Tutorials/1.9/friction})}
    \item[\textdagger\textdagger] \scriptsize{The average cumulative reward per episode was taken over 25 episodes.}
\end{tablenotes}
\end{sc}
\end{small}
\end{center}
\vskip -0.1in
\end{threeparttable}
}
    \subsection{Parameter optimization}
    
    For all three RL implementations, we used a discount factor of $\gamma = 0.9$.
    
	\subsubsection{Learning rates}\label{subsubsection:learning_rates}
    
    Determining optimal learning rates for these RL techniques is very important, as they can either make or break the algorithm's performance. 
    
	After testing Q-learning with different learning rates $\alpha$, and measuring their learning performance, we determined that learning rates between 0.01 and 0.2 work well for our purposes. In our analysis of Q-learning's performance in Section \ref{subsection:qlearning_analysis}, we use a learning rate of $\alpha = 0.1$.

    Since the action space $\{ (v_\ell, v_\omega)^\intercal \}$ for PG is 2-dimensional, the episodic REINFORCE algorithm inputs two learning rates $\alpha_\ell$ and $\alpha_\omega$, one for each of $v_\ell$ and $v_\omega$ respectively. Letting $U = \{ 10^{-j} : j \in \{3, 4, \ldots, 8 \} \}$, we chose learning rates for each task using the following method: we tested every pair $(\alpha_\ell, \alpha_\omega) \in U \times U$ and chose the pair that yielded the policy with the highest average cumulative reward per episode. Table \ref{table:pgtable} lists the pair of learning rates chosen for each task using this method.


\section{Testing and Analysis}
\label{testing_and_analysis}
    \subsection{Q-learning}\label{subsection:qlearning_analysis}
   	
    \subsubsection{User Policy}
    
    One drawback of the Q-learning algorithm is that the learning cost becomes prohibitively expensive as the state space grows large. To try to speed up the learning, we implemented a user policy $\pi_u: S \times A \mapsto [0, 1]$ that allows the robot to learn by demonstration. The idea is to use the user's trajectories to bootstrap the Q-table for learning. The user's trajectories are sub-optimal, so RL will refine them to create $\pi$.
    
    We first manually drive the robot around using keyboard teleop, creating an arbitrary number of user trajectories $\mathbb{T}_u$ that eventually lead the robot to the target object. Then these user trajectories are stored and used to compute the user policy $\pi_u$, where for every state-action pair $(s,a)$, $\pi_u(s, a)$ is the probability that the user chooses action $a$ in state $s$. When the robot follows the user policy $\pi_u$ in a state $s$, it computes the nearest state $s' \in \mathbb{T}_u$ to $s$, then chooses the most probable action $\argmax_a \pi_u(s', a)$.
    
    The robot initially starts out with numbers $p, q \in [0, 1]$, $p+q \leq 1$, such that for every time step $t$, the robot chooses a random action with probability $p$, follows the user policy with probability $q$, and chooses the best action based on its current policy $\pi_t$ otherwise. Over time, as the robot's policy $\pi_t$ becomes closer to optimal, $p$ and $q$ are slowly decreased until the robot can autonomously complete its task without the user's help. In our experiments, we decreased $p$ and $q$ by $1\%$ every 1,000 episodes.

\begin{figure}[t]
\vskip 0.2in
\begin{center}
\centerline{\includegraphics[width=\columnwidth]{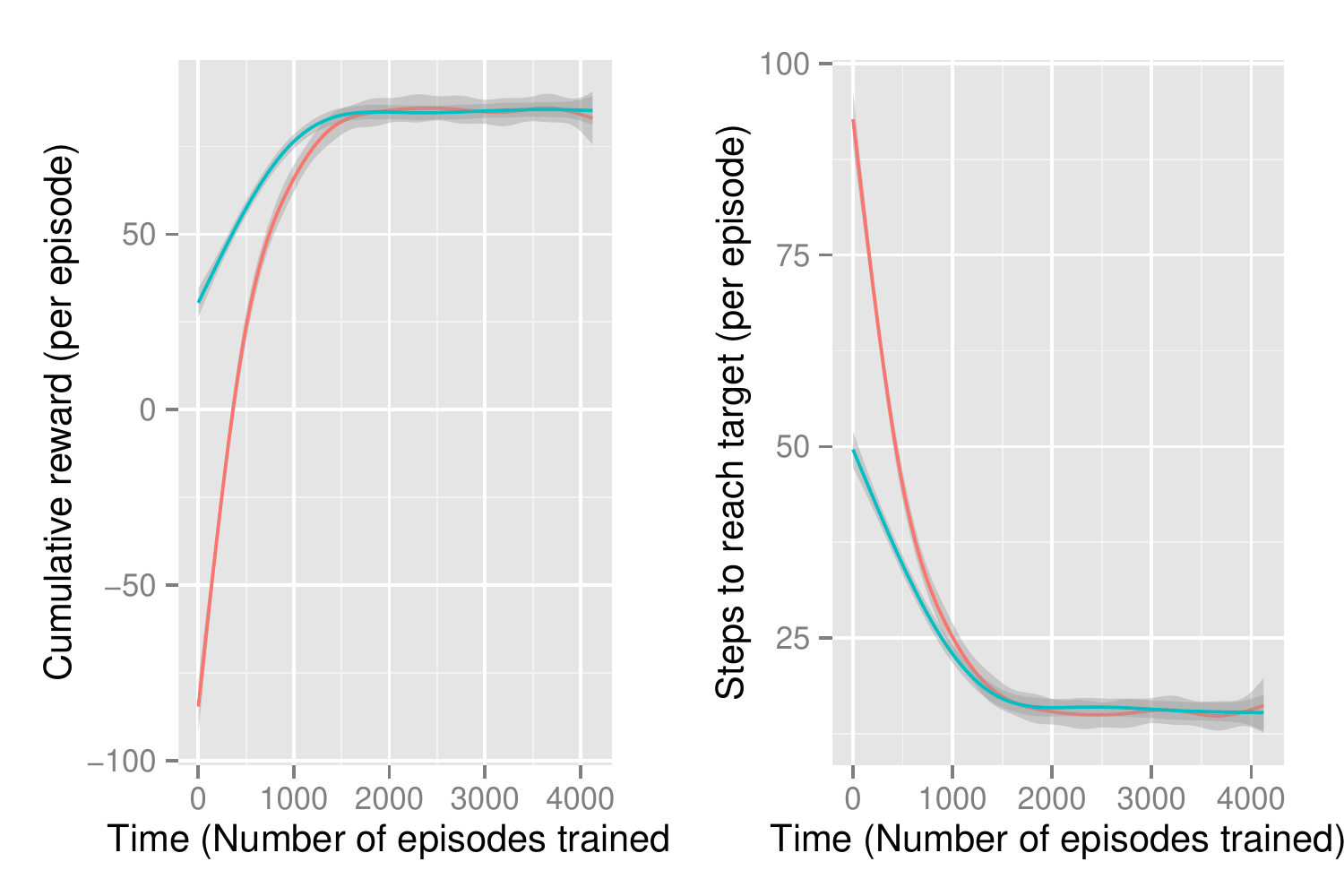}}
\caption{Comparison of Q-learning performance with (blue) and without (red) user policy. The plots in both graphs seem to converge after about 1,700 episodes, showing that the learning rates are about the same.\label{fig_qlearning}}
\end{center}
 \vskip -0.2in
\end{figure}

    In Fig. \ref{fig_qlearning}, we compare the results from two experiments I and II, where in I we initially set $p=.2$ and $q=.65$ (the robot having a relatively high probability of following $\pi_u$), and in II we initially set $p=.2$ and $q=0$ (no user policy). We ran both experiments for 4,000 episodes each, recording the cumulative reward and the total number of steps taken to reach the target in every episode. In both graphs the plots seem to converge after about 1,700 episodes, although the plots from experiment II have greater variance. This suggests that the learning rates for Q-learning with and without user policy are similar, although having user policy may more quickly yield a more stable policy.
    
	\subsubsection{Size of the state space}\label{subsubsection:state_space_size}
    
    The learning cost for Q-learning grows increasingly expensive as the size of the state space grows. To illustrate this effect, we also experimented Q-learning with a larger 3-dimensional state representation $\{ ([d], [\beta]/12, [\zeta]/12)^\intercal \}$, where $\beta$ (resp. $\zeta$) is the counterclockwise angle in degrees from the x-axis of the 2D map of the environment to $\bm{v}$ (resp. $\bm{u}$). (See Fig. \ref{fig_staterep} for an illustration.) We note that $\omega = \beta - \zeta$.

\begin{figure}[t]
\vskip 0.2in
\begin{center}
\centerline{\includegraphics[width=\columnwidth]{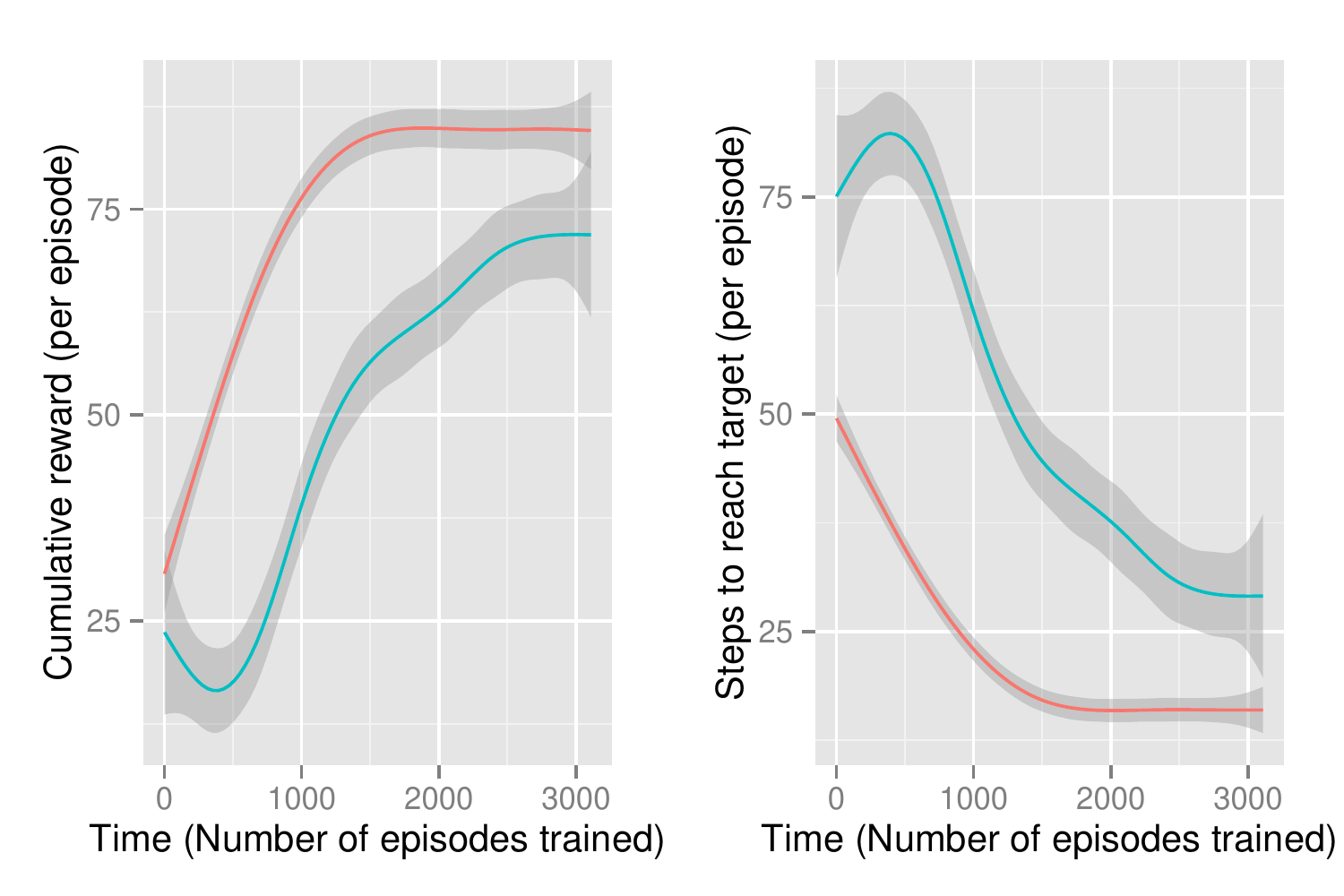}}
\caption{Comparison of Q-learning's performance with 2-dimensional (red) and 3-dimensional (blue) state spaces. In both plots, the agent's performance improves at a faster rate and achieves better results with the smaller state space. \label{fig:plotqlearning2vs3}}
\end{center}
 \vskip -0.2in
\end{figure}

In Fig. \ref{fig:plotqlearning2vs3}, we compare Q-learning's performance with the 2-dimensional state representation $\{ ([d], [\omega]/12)^\intercal \}$ and with the 3-dimensional state representation $\{ ([d], [\beta]/12, [\zeta]/12)^\intercal \}$. In both plots, the agent's performance improves at a faster rate and achieves significantly better results with the smaller state space. This illustrates the problem with Q-learning in a realistic environment, and shows where MTL could be especially beneficial.

    \subsection{Policy Gradient Reinforcement Learning}

For PG, we also used a ``user policy'' in order to quicken learning. More specifically, we experimentally determined a user policy $\pi_u$ with parameters $\bm{\theta}_u$ that (at least qualitatively) performs only slightly worse than optimal. To analyze the performance of PG, we initialized the parameters $\bm{\theta}$ of the robot's policy $\pi$ to zero and then trained the robot for 400 episodes, using $\pi_u$ to select its actions. Then for another 400 episodes, we used $\pi$ to select the robot's actions,  recording the cumulative reward and the number of steps to reach the target for every episode. Using these criteria, we found that $\pi$ performs at least as well, if not better, than the user policy $\pi_u$ (see Table \ref{table:pganalysis}).

\begin{table}[t]
\caption{A performance comparison between the policy $\pi$ trained using PG over 400 episodes and the user policy $\pi_u$. The average cumulative reward per episode and the average number of steps to reach the target for the two policies are very similar.\label{table:pganalysis}}
\vskip 0.15in
\begin{center}
\begin{small}
\begin{sc}
\begin{tabular}{lcccr}
\hline
\abovespace\belowspace
Policy & Avg. Cumul. Reward & Avg. \# of steps\tabularnewline \\
\hline
\abovespace
$\pi_u$& 92.394 & 16.198\tabularnewline \\
$\pi$ & 93.839 & 17.088\tabularnewline \\
\hline
\end{tabular}
\end{sc}
\end{small}
\end{center}
\vskip -0.1in
\end{table}

We note that the robot learned a policy almost as good as optimal in a shorter period of time than it did using Q-learning. PG has several advantages over Q-learning for this given task, as function approximation allows PG to handle continuous actions and states and even imperfect state information.

    \subsection{PG-ELLA}
    
    Due to the lack of time and resources, we have not done a complete empirical analysis of our PG-ELLA implementation. For instance, the ROS I/O Package could not be installed on the GRASP Lab workstation computers because the MATLAB software installed on these computers were outdated, and thus all of the experiments for both PG and PG-ELLA had to be run on a personal laptop. The limited experiments with PG-ELLA thus far were not as successful, primarily because many of PG-ELLA's assumptions were violated. For instance, we did not have enough tasks nor enough trajectories. We leave a more comprehensive empirical analysis of PG-ELLA to future work.


\section{Conclusion \& Future Works}
 
Because reinforcement learning can easily become prohibitively expensive, we worked with a simplified domain in the limited time we had for this project. But now the next step is to carry PG-ELLA into a more complex domain and do empirical analysis on its performance. We can add less informative features to the state vector, such as the robot's position $(x_A, y_A)$ on the map, or the robot's perception-based data (e.g., is the target visible?). We can also add obstacles to the environment, essentially turning the map into a maze. Additionally, we can have variable environment dynamics: for example, instead of having a uniform ground friction coefficient, we can place patches of rough carpet or sand on the ground. Lastly, we can move PG-ELLA onto a robot in the real world, where imperfect and noisy sensor data are common.

Of course, learning will quickly become more expensive, as we have demonstrated with the larger state space for Q-learning in Section \ref{subsubsection:state_space_size}. But we hope that through the sharing of knowledge between multiple tasks, the robot will be able to overcome the prohibitive cost of reinforcement learning.

\section{Acknowledgements}

This work was supported by the National Science Foundation REU Site at the GRASP Laboratory in the University of Pennsylvania. I would like to express my very great appreciation to Professor Eric Eaton, my research advisor, for his patient guidance and valuable suggestions during the planning and development of this research work. I would also like to express my sincere gratitude to my mentors Christopher Clingerman and Dr. Haitham Bou Ammar for their invaluable help and support during this research project.

\nocite{ReinforcementLearningI}
\bibliography{reu_paper}

\begin{thebibliography}{7}
\providecommand{\natexlab}[1]{#1}
\providecommand{\url}[1]{\texttt{#1}}
\expandafter\ifx\csname urlstyle\endcsname\relax
  \providecommand{\doi}[1]{doi: #1}\else
  \providecommand{\doi}{doi: \begingroup \urlstyle{rm}\Url}\fi

\bibitem[{Bou Ammar} et~al.(2014){Bou Ammar}, Eaton, Ruvolo, and
  Taylor]{pgella14}
{Bou Ammar}, H., Eaton, E., Ruvolo, P., and Taylor, M.~E.
\newblock Online multi-task learning for policy gradient methods.
\newblock \emph{International Conference on Machine Learning}, 2014.

\bibitem[Caruana(1997)]{caruana97}
Caruana, R.
\newblock \emph{Multitask Learning}.
\newblock PhD thesis, School of Computer Science, Carnegie Mellon University,
  1997.

\bibitem[Kober \& Peters(2008)Kober and Peters]{Kober08}
Kober, J. and Peters, J.
\newblock Policy search for motor primitives in robotics.
\newblock \emph{Advances in Neural Information Processing Systems (NIPS)},
  2008.

\bibitem[Ruvolo \& Eaton(2013)Ruvolo and Eaton]{ella13}
Ruvolo, P. and Eaton, E.
\newblock {ELLA}: An {E}fficient {L}ifelong {L}earning {A}lgorithm.
\newblock \emph{International Conference on Machine Learning}, 2013.

\bibitem[Sutton \& Barto(1998)Sutton and Barto]{ReinforcementLearningI}
Sutton, R.~S. and Barto, A.~G.
\newblock \emph{Reinforcement Learning: An Introduction}.
\newblock The MIT Press, Cambridge, MA, 1998.

\bibitem[Thrun(1996)]{Thrun96b}
Thrun, S.
\newblock \emph{Explanation-Based Neural Network Learning: A Lifelong Learning
  Approach}.
\newblock Kluwer Academic Publishers, Boston, MA, 1996.

\bibitem[Williams(1992)]{Williams92}
Williams, R.~J.
\newblock Simple statistical gradient-following algorithms for connectionist
  reinforcement learning.
\newblock \emph{Machine Learning}, 8:\penalty0 229--256, 1992.

\end{thebibliography}
\bibliographystyle{icml2014}

\end{document}